\pdfoutput=1
\documentclass{article}
\usepackage[final]{colm2026_conference}

\usepackage{microtype}
\usepackage{hyperref}
\usepackage{url}
\usepackage{booktabs}
\usepackage{amsmath}
\usepackage{amssymb}
\usepackage{graphicx}
\usepackage{algorithm}
\usepackage{algorithmic}
\usepackage{multirow}
\usepackage{xcolor}
\usepackage{subcaption}
\usepackage{tabularx}
\usepackage{lineno}
\usepackage{float}
\usepackage{placeins}
\usepackage{needspace}
\usepackage[most]{tcolorbox}
\usepackage{listings}

\definecolor{pbClassify}{HTML}{4A5B6E}   
\definecolor{pbQuery}{HTML}{2C6E8A}      
\definecolor{pbVerify}{HTML}{2E6B4F}     
\definecolor{pbJudge}{HTML}{7A4A63}      
\definecolor{pbSlot}{HTML}{B4551F}       
\definecolor{pbBody}{HTML}{1A1A1A}

\lstdefinestyle{pbstyle}{
  basicstyle=\footnotesize\ttfamily\color{pbBody},
  breaklines=true,
  breakautoindent=false,
  columns=fullflexible,
  keepspaces=true,
  showstringspaces=false,
  aboveskip=0pt, belowskip=0pt,
  moredelim=[s][\color{pbSlot}\bfseries]{\{}{\}},
  escapeinside={(*@}{@*)},
}

\tcbset{
  pbbase/.style={
    enhanced, breakable,
    listing only, listing options={style=pbstyle},
    boxrule=0pt, frame hidden,
    arc=2pt, outer arc=2pt,
    left=7pt, right=7pt, top=6pt, bottom=6pt,
    toptitle=2.5pt, bottomtitle=2.5pt,
    coltitle=white, fonttitle=\footnotesize\bfseries,
    borderline west={2pt}{0pt}{tcbcolbacktitle},
    attach boxed title to top left={xshift=0mm, yshift=0mm},
    boxed title style={sharp corners=downhill, arc=2pt, boxrule=0pt},
  },
  pbaccent/.style 2 args={pbbase, colbacktitle=#1, colback=#1!4!white},
}

\NewTCBListing[auto counter, number within=section]
  {promptbox}{O{pbQuery} m o}{%
    pbaccent={#1}{},
    title={Prompt~\thetcbcounter\quad #2},
    IfValueT={#3}{label={#3}},
  }
\definecolor{darkblue}{rgb}{0, 0, 0.5}
\hypersetup{colorlinks=true, citecolor=darkblue, linkcolor=darkblue, urlcolor=darkblue}

\title{CABLE: Extending the Reach of Memory Retrieval via \\
Complementary Antecedent-Based Linking and Expansion}

\author{
Zheling Tan$^{1}$, Jin Gao$^{1}$, Dequan Wang$^{1,2}$
\thanks{Corresponding author: \texttt{dequanwang@sjtu.edu.cn}}\\
$^{1}$Shanghai Jiao Tong University
$^{2}$Shanghai Innovation Institute
}

\begin{document}

\ifcolmsubmission
\linenumbers
\fi

\maketitle

\begin{abstract}
As LLM agents operate across structured workflows and sessions, preserving long-term history does not ensure that later contexts can recover relevant evidence through a bounded memory interface. We study this evidence-reachability problem in long-term conversational memory, where retrieval still relies heavily on semantic similarity. This works well for topical recall, but it often misses
earlier experiences, plans, or motivations that are semantically distant from
the later events they help explain.
Existing memory graphs provide cross-memory structure, yet links driven mainly
by semantic overlap can duplicate what the host retriever already recovers.
We argue that link construction should instead prioritize a sparse set of retriever-complementary associations.
We present CABLE (\textbf{C}omplementary
\textbf{A}ntecedent-\textbf{B}ased
\textbf{L}inking and \textbf{E}xpansion), a plug-in augmentation that constructs  links designed to extend the host retriever's direct semantic reach.
For each new memory, CABLE generates antecedent-oriented queries, retrieves
prior memories, subtracts candidates in the direct semantic neighborhood, and
verifies the remainder before adding the accepted complementary associations
into a sparse directed graph.
At retrieval time, CABLE expands the host system's retrieved seeds along these
links to surface implicit supporting evidence.
We evaluate CABLE with A-MEM on LoCoMo and MA-LongMemEval, and further
integrate it into SimpleMem and Mem0\textsuperscript{g} on LoCoMo, using
Qwen3.5-27B, DeepSeek-chat, and GPT-4o-mini. CABLE yields higher mean LLM-judge scores
in every evaluated system-level setting, with the largest gains in categories
where useful evidence is distributed across memories or sessions, including
open-domain, multi-session, and preference-oriented questions.
These results support prioritizing sparse, reasoning-relevant associations that
complement rather than duplicate the host retriever.
The core CABLE implementation is available at
\url{https://github.com/TanZheling/CABLE}.
\end{abstract}

\section{Introduction}
\label{sec:intro}
LLM agent systems are increasingly organized as execution graphs of specialized
components and explicit state transitions~\citep{wu2024stateflow,
zhuge2024gptswarm, zhang2025aflow}. As these systems operate across component and session boundaries, long-term history can no longer be assumed to remain inside one prompt.
This creates an information-access problem: execution graphs specify where computation proceeds, but not how later components can recover the relevant history within a bounded context.

\begin{figure}[t]
    \centering
    \includegraphics[width=\linewidth]{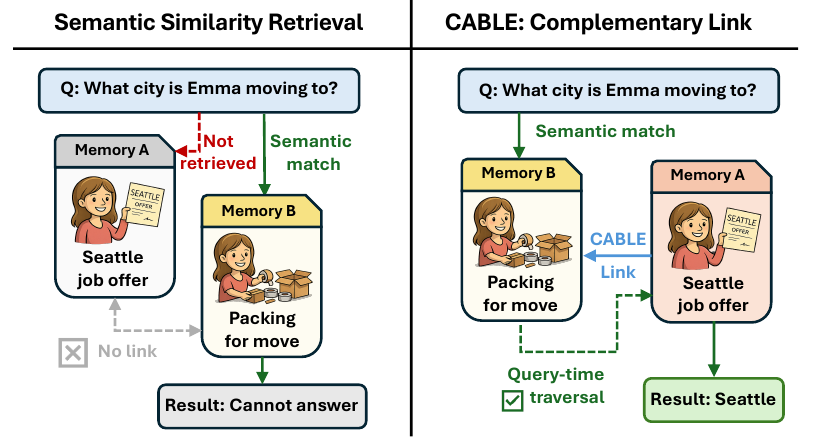}
    \caption{ Stored evidence can remain outside direct retrieval. Direct semantic retrieval returns Memory B but not Memory A, which contains the required answer. During construction, CABLE stores the antecedent link ($A\rightarrow B$); at query time, expansion starts from the retrieved Memory B and follows its incoming link to recover Memory A.}
    \label{fig:banner}
\end{figure}

Long-term conversational memory provides a controlled setting for studying one concrete instance of this evidence-reachability problem: whether stored evidence can still be retrieved when it lies outside the direct semantic neighborhood of a later query. Existing systems typically store past interactions as
compact memory entries and retrieve a small subset for each new
query~\citep{park2023generative,packer2023memgpt,hu2025memory}. Semantic
retrieval is effective when the query and the required memory are close in
embedding space, even if their surface forms differ. It is less reliable when
the query concerns a later event but answering it requires an earlier memory
about an experience, plan, motivation, or background event. In such cases, the required evidence is present in memory but inaccessible within the retriever’s limited output budget.
Figure~\ref{fig:banner} illustrates this distinction. The question ``What city
is Emma moving to?'' retrieves a recent memory about packing, while the answer
appears in an earlier memory about a Seattle job offer. The first memory is
topically close to the query; the second supplies the missing antecedent.

Structured memory offers a natural way to extend direct retrieval. Mem0
extracts entity-level relations~\citep{chhikara2025mem0}, A-MEM links memories
through contextual descriptions~\citep{xu2025mem}, and HippoRAG reasons over an
LLM-constructed knowledge graph~\citep{gutierrez2024hipporag}. Yet adding links
does not necessarily add access. When associations mainly reflect
semantic overlap, shared entities, or nearby descriptions, graph expansion can
return memories already within the host retriever's reach. Under a fixed
context budget, redundant neighbors can displace more useful evidence without
extending the system's effective reach.

This observation suggests a design criterion for memory graphs: associations should be \emph{retriever-complementary}. Their value lies in exposing useful evidence beyond the host retriever's direct neighborhood. Such links
need not form a comprehensive world model; they can remain sparse, persistent,
and inspectable while providing alternative paths to semantically
distant memories. This shifts the role of structured memory from representing
relations for their own sake to providing marginal retrieval value under
bounded context.

We instantiate this principle with CABLE (\textbf{C}omplementary
\textbf{A}ntecedent-\textbf{B}ased \textbf{L}inking and
\textbf{E}xpansion), a plug-in augmentation for existing memory systems. When
a new memory is written, CABLE asks what earlier experiences, plans,
motivations, or background events might have led to it and converts these
hypotheses into antecedent-oriented queries. It retrieves the host system's
direct semantic neighborhood and candidate antecedents, subtracts their
overlap, and verifies the remaining candidates before storing directed links.
At query time, CABLE expands the host retriever's seed memories by one hop,
aggregates support from connected seeds, and filters redundant additions.
In the A-MEM and Mem0\textsuperscript{g} integrations, CABLE operates under a fixed retrieved-entry budget, replacing lower-ranked baseline memories rather than increasing the number of entries passed to the answer generator.
Antecedent reasoning is performed once during memory construction, and the resulting links are reused across subsequent retrievals. Retrieval-time expansion therefore requires no additional LLM calls.

CABLE targets one capability that a durable memory layer should provide:
stored history should remain recoverable when later queries require indirectly related evidence outside their direct retrieval neighborhood.
We isolate this capability through long-term
conversational memory question answering, without introducing tool use,
multi-agent orchestration, or graph-level task completion.

We evaluate CABLE with A-MEM on LoCoMo and MA-LongMemEval, and further integrate it into SimpleMem and graph-enabled Mem0\textsuperscript{g} on LoCoMo, using Qwen3.5-27B, DeepSeek-chat, and GPT-4o-mini. CABLE yields higher overall mean LLM-judge scores in all evaluated system-level settings. The gains are largest on categories that require implicit cross-memory association, including open-domain questions in LoCoMo and multi-session or preference-oriented questions in MA-LongMemEval.

Our contributions are threefold:
\begin{itemize}
\item We formulate \emph{retriever complementarity} as a design principle
for structured memory: useful associations extend rather than duplicate
the host retriever's effective reach.
\item We propose CABLE, which constructs sparse antecedent links through
dual retrieval, overlap subtraction, and verification, then reuses them
through bounded expansion without retrieval-time LLM calls.
\item We demonstrate consistent overall improvements across two long-term
conversational memory benchmarks, three structurally different memory
systems, and multiple LLM backbones under controlled retrieval protocols.
\end{itemize}

\section{Related work}
\label{sec:related}

\paragraph{Entry-based memory systems.}
To retain information beyond the context window, recent work has introduced explicit memory modules for LLM agents that store and manage past interactions as discrete memory entries. These systems differ primarily in how such entries are formed and maintained. MemoryBank~\citep{zhong2024memorybank} treats long-term memory as an evolving collection of user-related records, updating them with a forgetting-aware mechanism to model temporal decay. Mem0~\citep{chhikara2025mem0} extracts salient facts from ongoing conversations, consolidates them into persistent memory items, and retrieves them when needed for downstream generation. LightMem~\citep{fang2025lightmem} improves efficiency through lightweight compression, topic-based grouping, and offline consolidation, turning raw interaction history into compact memory units organized for structured access. SimpleMem~\citep{liu2026simplemem} likewise emphasizes compact entry construction, distilling interactions into multi-view indexed memory units and recursively consolidating related units into higher-level abstractions. MemInsight~\citep{salama2025meminsight} further augments stored interactions to improve the semantic representation and retrieval quality of memory entries.
Despite their differences in implementation, these systems primarily improve how individual memory entries are extracted, compressed, updated, and retrieved.
Across these designs, downstream reasoning still receives only a selected set of entries; evidence outside it requires an alternative access path across memories.

\paragraph{Structured memory association.}

Beyond storing memories as largely independent entries, recent work has explored how stronger relational structure can improve associative access in memory systems. A-MEM~\citep{xu2025mem} enables memories to evolve through dynamic indexing and linking, forming a self-organized network of related notes.
This line of work echoes linked note-taking and associative indexing~\citep{bush1945think,ahrens2017smart}, which use persistent cross-references to provide access paths beyond storage and search alone.
Mem0 further offers a graph-based variant, Mem0\textsuperscript{g}~\citep{chhikara2025mem0}, that represents memories as a directed labeled graph in which entities serve as nodes and their relationships as edges, enabling multi-hop traversal across related facts.
CompassMem~\citep{hu2026memory} moves to an event-centric design, segmenting experience into events and connecting them through explicit logical relations, so that agents can navigate memory through an Event Graph rather than rely solely on direct item-level matching. Other methods further couple enriched links with reasoning-time control. ActMem~\citep{zhang2026actmem} constructs a causal and semantic graph and integrates retrieval with counterfactual reasoning, while MAGMA~\citep{jiang2026magma} represents each memory item through multiple relational views and performs policy-guided traversal over them. Hindsight~\citep{latimer2025hindsight} likewise treats memory as a structured substrate, supporting temporal and entity-aware recall over organized memory networks.
Recent systems also explore temporally grounded property graphs, query-adaptive multi-relational graph retrieval, and hierarchical evidence selection~\citep{banerjee2026apex,van2026memorai,cao2026higmem}.
These methods enrich memory representation and retrieval, but do not explicitly prioritize links by the additional evidence they provide beyond the host retriever. Under a bounded context budget, redundant neighbors add structure without extending effective reach.

\paragraph{Retrieval beyond direct matching.}
Retrieval is a central component of memory systems, as it determines how stored information can be brought back into the reasoning process. Rather than relying only on direct matching between the input query and stored text, a number of methods improve retrieval by constructing intermediate query-side representations. HyDE~\citep{gao2023precise} is a representative example: it first generates a hypothetical document from the query and retrieves evidence using the embedding of that synthetic text. HyPE~\citep{vake2025bridging} shifts this idea to the indexing stage by precomputing question-like prompts for each chunk. Question Decomposition for RAG~\citep{ammann2025question} improves retrieval for complex questions by decomposing them into sub-questions and retrieving evidence for each part, while GenGround~\citep{shi2024generate} interleaves generation and retrieval through intermediate question-answer pairs.
These methods show that alternative query representations can broaden
candidate discovery at inference or indexing time. CABLE likewise uses generated queries for candidate discovery, but subtracts the direct retrieval set, verifies the remainder, and stores accepted links for reuse.

\section{Method}
\label{sec:method}

\begin{figure}[t]
    \centering
    \includegraphics[width=\textwidth]{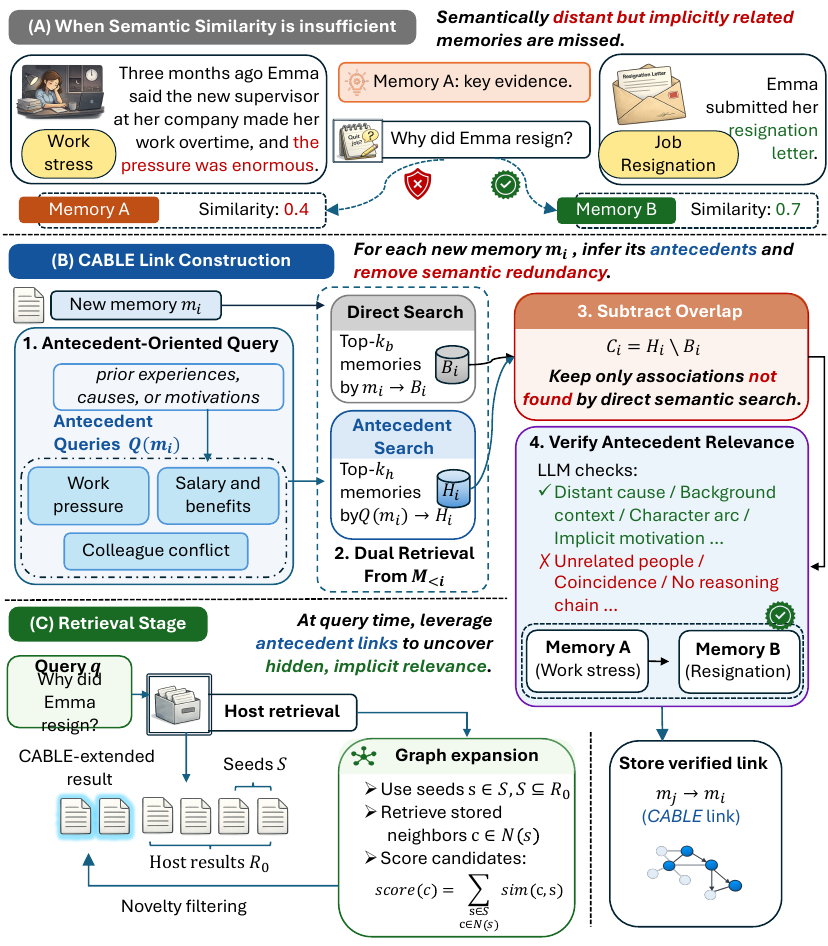}
    \caption{CABLE constructs sparse antecedent links and reuses them for bounded retrieval expansion. (A) Direct semantic retrieval reaches Memory B but misses the semantically distant yet relevant Memory A. (B) During construction, CABLE retrieves direct matches ($B_i$) and antecedent candidates ($H_i$), subtracts their overlap, verifies the remaining candidates, and stores accepted links ($m_j\rightarrow m_i$) in graph G. (C) At inference, the host system returns ($R_0$), CABLE selects reliable seeds ($S$), expands their stored graph neighborhoods, and applies candidate scoring and novelty filtering. Similarity values are illustrative.}
\label{fig:method}
\end{figure}

Figure~\ref{fig:method} summarizes CABLE. During memory construction (panel B), CABLE generates antecedent-oriented queries for each new memory, retrieves both direct semantic neighbors and antecedent candidates, removes the overlap, and verifies the remaining candidates before adding a sparse set of directed edges. At query time (panel C), CABLE expands the host system's retrieved seeds along these edges and filters redundant additions. Together, these steps aim to introduce only non-redundant links that provide additional retrieval value. Throughout, $G=(M,E)$ denotes the directed graph CABLE maintains over all memory entries, and $\phi(\cdot)$ the embedding function of the host system's retriever.

\subsection{Construction-time antecedent link building}
Let $M_{<i} = \{m_1, \ldots, m_{i-1}\}$ denote the memory base when a new memory $m_i$ arrives. CABLE builds a sparse set of directed edges from prior memories in $M_{<i}$ to $m_i$.
These edges provide alternative access paths
to prior memories that the host system's semantic retriever is unlikely to
surface directly.

\subsubsection{Antecedent-oriented query generation}
\label{sec:causal_reasoning}

CABLE broadens candidate discovery by converting hypotheses about earlier
experiences, plans, motivations, or background events into antecedent-oriented
queries. Overlap subtraction and verification then determine which candidates
become stored links.

Before generation, a rule-based filter removes low-information memories such as greetings. An LLM then classifies each remaining memory into an operational type such as \texttt{event}, \texttt{opinion}, \texttt{plan}, or \texttt{state\_change}. Guided by this type, it generates a small set of antecedent-oriented queries $Q(m_i)=\{q_1,\ldots,q_{N_q}\}$. For an \texttt{event}, for example, the queries can emphasize preceding events, related plans, motivations, or prior experiences. In panel B of Figure~\ref{fig:method}, the memory \emph{``Emma submitted her resignation letter''} yields queries such as \emph{``work pressure,''} \emph{``colleague conflict,''} and \emph{``salary and benefits.''}

\subsubsection{Dual retrieval}
\label{sec:dual_retrieval}
CABLE next performs two parallel retrievals over the memory base $M_{<i}$.

\begin{itemize}
    \item \textbf{Direct search.} $B_i$ contains the top-$K_b$ memories whose embeddings have the highest cosine similarity to $m_i$. This set serves as the construction-time direct semantic neighborhood around $m_i$.
    \item \textbf{Antecedent search.} Each antecedent query $q \in Q(m_i)$ is issued separately, and $H_i$ is the union of the top-$K_h$ memories returned for each query. If the same memory is returned by more than one query, we keep its highest score.
\end{itemize}

\subsubsection{Overlap subtraction}
\label{sec:subtract}

Given the two retrieved sets, CABLE keeps only the retriever-complementary candidates:
\[
C_i = H_i \setminus B_i.
\]
Any memory already included in the top-$K_b$ direct semantic retrieval set is excluded from the complementary candidate set $C_i$. If $C_i=\emptyset$, no CABLE edge is added for $m_i$.

\subsubsection{Verification and graph update}
\label{sec:verify}

Each candidate $m_j \in C_i$ is then verified by an LLM.
The verifier accepts candidates that provide useful prior context, such as a
cause, motivation, enabling event, background event, or earlier state, and
rejects topical co-occurrence or entity overlap alone.
This verification step prevents weak or coincidental associations from unnecessarily densifying the graph. Each accepted pair adds a directed edge
\[
E \leftarrow E \cup \{(m_j \rightarrow m_i)\},
\]
where $m_j \rightarrow m_i$ denotes a verified antecedent link from $m_j$ to $m_i$.

\subsection{Retrieval-stage extension}
\label{sec:retrieval}

At query time, the host memory system first executes its standard retrieval to produce an initial result set $R_0$. CABLE then expands $R_0$ through the graph $G$ using seed selection, candidate scoring, and novelty filtering.

\subsubsection{Seed selection and candidate scoring}
\label{sec:scoring}
Given query $q$, we keep only reliable seeds
\[
S = \{s \in R_0 \mid \mathrm{sim}(\phi(q), \phi(s)) \ge \tau\},
\]
where $\mathrm{sim}(\cdot,\cdot)$ is cosine similarity and $\tau$ is a seed-quality threshold. For each seed $s$, let $N(s)$ be its one-hop neighbors in $G$, including both incoming and outgoing edges. Incoming edges surface earlier memories that help explain a retrieved event, while outgoing edges surface later memories that a retrieved cause or plan helps explain. The candidate pool is
\[
U = \Big(\bigcup_{s \in S} N(s)\Big) \setminus R_0.
\]
We score each candidate $c \in U$ by aggregating support from connected seeds:
\[
\mathrm{score}(c) = \sum_{\substack{s \in S \\ c \in N(s)}} \mathrm{sim}(\phi(c), \phi(s)).
\]
This favors candidates supported by multiple high-confidence seeds.

\subsubsection{Novelty filtering}
\label{sec:novelty}
Candidates are ranked by $\mathrm{score}(c)$ and selected greedily up to the
expansion budget. To avoid reintroducing near-duplicates, CABLE accepts a candidate only if
\[
\max_{r \in R} \mathrm{sim}(\phi(c), \phi(r)) < \theta,
\]

where $R$ is the current result set and $\theta$ is a novelty threshold. The output thus preserves the host system's retrieval backbone while adding only non-redundant graph expansions.
Construction cost, graph-growth bounds, and retrieval-time overhead are
discussed in Appendix~\ref{app:overhead}.

\section{Experiments}
\label{sec:experiments}

\subsection{Experimental setup}
\label{sec:setup}

\subsubsection{Benchmarks}
\label{sec:benchmarks}

We evaluate CABLE on two long-term memory dialogue benchmarks.

\textbf{LoCoMo}~\citep{maharana2024evaluating} evaluates memory and reasoning over long conversations. We exclude the adversarial split and report the four answerable categories: \textit{Single-hop} (841), \textit{Multi-hop} (282), \textit{Temporal Reasoning} (321), and \textit{Open-domain Knowledge} (96), for a total of 1,540 questions. These categories respectively emphasize direct retrieval, multi-memory composition, temporal dependencies, and broader user- or world-aware reasoning.

\textbf{MA-LongMemEval}~\citep{wu2024longmemeval} evaluates long-term memory in multi-session dialogues. We use the reformulated setting from MemoryAgentBench~\citep{hu2025evaluating}, which reorganizes the original benchmark into five long dialogue contexts paired with 300 questions. We report six non-abstention question types: \textit{multi-session}, \textit{temporal-reasoning}, \textit{knowledge-update}, \textit{single-session-user}, \textit{single-session-assistant}, and \textit{single-session-preference}.

\subsubsection{Memory systems}
\label{sec:memory_systems}

We integrate CABLE into three representative memory systems.

\textbf{A-MEM}~\citep{xu2025mem} is an agentic memory system with dynamic organization, explicit inter-memory links, and continuous memory evolution. We use it to test whether CABLE remains useful even when the host already supports memory association. For a controlled comparison, both the baseline and CABLE pass at most 45 memory entries to the answer generator. CABLE replaces up to five of the lowest-ranked baseline entries rather than appending additional context.

\textbf{SimpleMem}~\citep{liu2026simplemem} is a recent memory framework that distills raw dialogues into compact, context-independent memory units via semantic structured compression, indexes each unit across semantic, lexical, and symbolic layers, and retrieves them through intent-aware planning with a hybrid scoring function and reflection-based sufficiency checking. This multi-stage pipeline makes it a strong baseline with sophisticated memory construction and retrieval.
Unlike A-MEM, SimpleMem uses host-controlled adaptive retrieval rather than a fixed retrieved-entry budget: its reflection step adaptively determines whether the retrieved memory context is sufficient.
We therefore activate CABLE only when this step judges the baseline retrieval insufficient, rather than under a fixed replacement budget.

\textbf{Mem0\textsuperscript{g}}
~\citep{chhikara2025mem0} is the graph-enabled variant of Mem0. It extracts entities and relations from memories and performs retrieval over an explicit directed knowledge graph. We enable its full graph-memory module to examine whether CABLE remains complementary to an existing graph-based memory system. For a controlled comparison, both settings retrieve 20 memories in total, with CABLE replacing up to five memories.

\subsubsection{Models}
\label{sec:models}
Within each experimental setting, we use the same backbone LLM throughout the main pipeline, including memory extraction, CABLE link construction, host system retrieval-time reasoning, and answer generation. On LoCoMo we evaluate Qwen3.5-27B~\citep{qwen35} and DeepSeek-V3.2-chat~\citep{liu2025deepseek}; on MA-LongMemEval we evaluate Qwen3.5-27B and GPT-4o-mini~\citep{hurst2024gpt}, the latter to test whether CABLE transfers to a closed-source model. The Mem0\textsuperscript{g} experiments use Qwen3.5-27B only. We refer to DeepSeek-V3.2-chat as DeepSeek-chat in the tables. The same backbone also serves as the LLM judge within each setting.

\subsubsection{Evaluation metrics}
\label{sec:evaluation_protocol}
We report the \textbf{mean LLM-judge score}. For each question, the judge assesses the semantic correctness of the generated answer relative to the reference and returns a score together with a short justification. We retain the returned score and average it across questions. The full judge prompt is provided in Appendix~\ref{app:prompt_judge}. We use this metric because answers on both benchmarks admit substantial lexical variation, making token-overlap measures less reliable as a primary evaluation signal. Within each benchmark-model setting, the baseline and +CABLE systems are evaluated using the same judge prompt and the same judge model.

\subsubsection{Implementation details}
\label{sec:implementation_details}
For link construction, CABLE generates at most $N_q=3$ antecedent-oriented queries per memory, and both the direct and antecedent searches retrieve $K_b=K_h=15$ candidates from $M_{<i}$. At retrieval time, CABLE uses an expansion budget of 5, seed-quality threshold $\tau=0.3$, and novelty threshold $\theta=0.9$. A-MEM uses all-MiniLM-L6-v2~\citep{wang2020minilm} as its embedding model, while SimpleMem uses Qwen3-Embedding-0.6B~\citep{qwen3embedding}, consistent with its original design. Mem0\textsuperscript{g} also uses Qwen3-Embedding-0.6B as its embedding model. Per-system integration protocols are detailed in Appendix~\ref{app:protocols}.

\subsection{Results on A-MEM}
\label{sec:amem_primary}

We first evaluate CABLE on \textbf{A-MEM} across \textbf{two benchmarks} and \textbf{multiple backbones}. We report overall results first and then analyze which question types benefit most.

\subsubsection{Overall results}
\label{sec:amem_overall}

\begin{table}[t]
\centering
\renewcommand{\arraystretch}{1.15}
\begin{tabular}{llrrr}
\toprule
\textbf{Benchmark} & \textbf{Model} &
\textbf{Baseline} & \textbf{+CABLE} &
\textbf{$\Delta$Score (pp)} \\
\midrule

LoCoMo
& Qwen3.5-27B
& 71.23
& \textbf{74.81}
& +3.58
\\

& DeepSeek-chat
& 68.15
& \textbf{70.26}
& +2.11
\\

\midrule

MA-LongMemEval
& Qwen3.5-27B
& 59.33
& \textbf{65.33}
& +6.00
\\

& GPT-4o-mini
& 48.67
& \textbf{49.67}
& +1.00
\\

\bottomrule
\end{tabular}
\caption{CABLE improves A-MEM in all four settings, with gains ranging from 1.00 to 6.00 percentage points. Scores are mean LLM-judge percentages; $\Delta$Score denotes +CABLE minus the matched baseline.}
\label{tab:amem_overall}
\end{table}

Table~\ref{tab:amem_overall} shows that CABLE improves A-MEM in all four evaluated settings. The largest system-level gain appears on MA-LongMemEval with Qwen3.5-27B (+6.00), while the gains on LoCoMo are +3.58 with Qwen3.5-27B and +2.11 with DeepSeek-chat. Even under GPT-4o-mini, CABLE remains beneficial (+1.00).

\subsubsection{Where does CABLE help?}
\label{sec:category_analysis}
\begin{table}[t]
\centering
\renewcommand{\arraystretch}{1.12}
\begin{tabular}{llrrr}
\toprule
\textbf{Backbone} & \textbf{Category} &
\textbf{Baseline} & \textbf{+CABLE} &
\textbf{$\Delta$Score (pp)} \\
\midrule

Qwen3.5-27B
& Single-hop
& 74.08
& \textbf{76.81}
& +2.73
\\

& Multi-hop
& 73.76
& \textbf{76.95}
& +3.19
\\

& Temporal
& 68.85
& \textbf{74.14}
& +5.29
\\

& Open-domain
& 46.88
& \textbf{53.12}
& +6.24
\\

\midrule

DeepSeek-chat
& Single-hop
& 73.96
& \textbf{75.51}
& +1.55
\\

& Multi-hop
& 63.65
& \textbf{65.25}
& +1.60
\\

& Temporal
& 64.49
& \textbf{66.36}
& +1.87
\\

& Open-domain
& 42.71
& \textbf{52.08}
& +9.37
\\

\bottomrule
\end{tabular}
\caption{CABLE improves every LoCoMo category under both backbones, with the largest gains on open-domain questions. Values are mean LLM-judge scores (\%).}
\label{tab:locomo_category}
\end{table}

On LoCoMo, Table~\ref{tab:locomo_category} shows gains in all categories under both models. The gains are relatively small on \textit{Single-hop}, where direct semantic retrieval is often sufficient, and consistently larger on \textit{Open-domain}, a pattern consistent with useful evidence lying outside the query's direct retrieval neighborhood. Under Qwen3.5-27B, gains are +2.73 on \textit{Single-hop} and +6.24 on \textit{Open-domain}; under DeepSeek-chat, \textit{Open-domain} has the largest gain (+9.37).

\begin{figure}[t]
    \centering
    \includegraphics[width=1.0\textwidth]{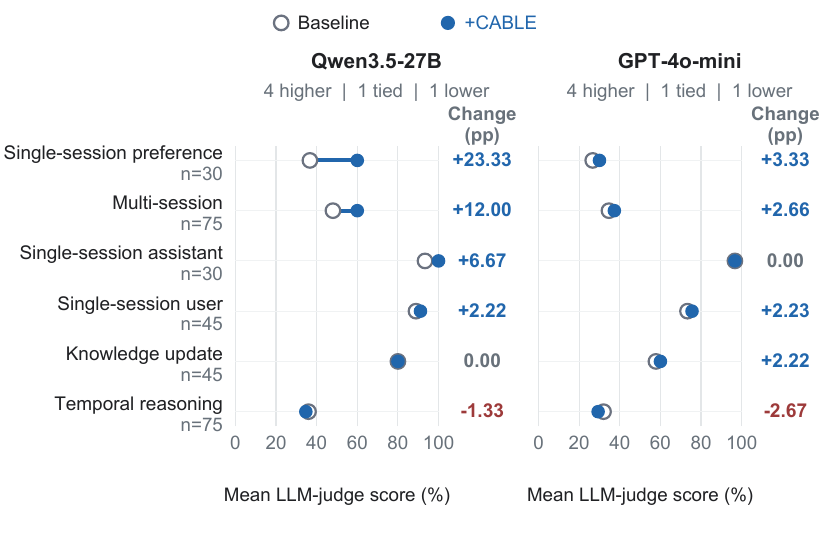}
    \caption{Category-wise A-MEM results on MA-LongMemEval. CABLE improves multi-session and single-session-preference under both backbones, but lowers temporal-reasoning. Labels report changes from baseline; higher is better.}
    \label{fig:lme_bar}
\end{figure}

As shown in Figure~\ref{fig:lme_bar}, on MA-LongMemEval, the strongest and most consistent gains appear in \textit{multi-session} and \textit{single-session-preference}. Under Qwen3.5-27B, CABLE improves \textit{multi-session} by +12.00 and \textit{single-session-preference} by +23.33; under GPT-4o-mini, the corresponding gains are +2.66 and +3.33. This pattern is consistent with CABLE helping when evidence is distributed
across sessions or encoded as preference cues. By contrast, \textit{temporal-reasoning} drops under both models (-1.33 and -2.67), suggesting that associative expansion is less helpful when the task depends on precise temporal resolution.

\subsection{Generalization across memory systems}
\label{sec:other_systems}
To examine whether CABLE depends on the design of A-MEM, we further integrate it into SimpleMem and Mem0\textsuperscript{g}, which represent adaptive hybrid retrieval and explicit graph-based retrieval, respectively. Table~\ref{tab:other_systems} summarizes the overall results.

\paragraph{Integration with SimpleMem.}
SimpleMem employs intent-aware retrieval planning with multi-view hybrid search across semantic, lexical, and symbolic indexes, followed by a reflection step that checks retrieval sufficiency. We activate CABLE retrieval only when the reflection step judges the baseline retrieval insufficient, so its role is explicitly complementary rather than always-on.

As shown in Table~\ref{tab:other_systems}, CABLE remains helpful on SimpleMem, with gains of +0.58 and +1.62. The absolute improvements are smaller than on A-MEM, which is consistent with SimpleMem's stronger reflection-based baseline and CABLE's selective activation.

\begin{table}[t]
\centering
\renewcommand{\arraystretch}{1.15}
\begin{tabular}{llrrr}
\toprule
\textbf{System} & \textbf{Model} &
\textbf{Baseline} & \textbf{+CABLE} &
\textbf{$\Delta$Score (pp)} \\
\midrule

SimpleMem
& Qwen3.5-27B
& 81.69
& \textbf{82.27}
& +0.58
\\

& DeepSeek-chat
& 79.42
& \textbf{81.04}
& +1.62
\\

\midrule

Mem0$^{g}$
& Qwen3.5-27B
& 52.60
& \textbf{54.80}
& +2.20
\\

\bottomrule
\end{tabular}
\caption{CABLE yields higher mean scores for adaptive-retrieval SimpleMem and graph-based Mem0g on LoCoMo, indicating that retriever complementarity is not specific to A-MEM. Comparisons are made within rows; $\Delta$Score is reported in percentage points.}
\label{tab:other_systems}
\end{table}

\paragraph{Integration with Mem0\textsuperscript{g}.}
CABLE improves Mem0\textsuperscript{g} from 52.6\% to 54.8\% under Qwen3.5-27B. Improvements are observed across all LoCoMo categories, including single-hop (+2.0), multi-hop (+2.5), temporal (+1.9), and open-domain (+4.2) questions.
CABLE improves Mem0\textsuperscript{g} while keeping the retrieved-entry count fixed, suggesting that its links can complement the host graph.

\subsection{Impact of CABLE Components}

Ablations in Table~\ref{tab:ablation} of Appendix~\ref{app:ablations} show that both overlap subtraction and LLM verification contribute to CABLE. Without overlap subtraction, semantically redundant links can consume the limited expansion budget without providing complementary evidence. Without verification, noisier candidate associations are retained, producing the largest degradation on MA-LongMemEval. Type-conditioned antecedent queries also outperform generic query decomposition by 0.46 percentage points on LoCoMo, indicating a modest benefit beyond topical query generation.

\subsection{Qualitative case study}
\label{sec:case_study}

Table~\ref{tab:case_study} in Appendix~\ref{app:case_study} illustrates how CABLE complements semantic retrieval on a LoCoMo example. The question asks why Melanie chose to use colors and patterns in her pottery. The baseline retriever returns \texttt{D5:6}, a memory about enjoying pottery in general, which is topically relevant but contains no information about her motivation. The query and \texttt{D5:6} are close in embedding space because they share surface-level topic words (e.g., pottery), yet topical proximity does not entail informational relevance.

CABLE reaches \texttt{D12:6} by following a stored antecedent link from \texttt{D5:6}, forming an elaborative bridge from the activity to the motivation behind it. The two memories discuss the same project yet use largely disjoint vocabulary: \texttt{D5:6} frames pottery as a relaxing hobby, while \texttt{D12:6} supplies the reference motivation: catching the eye and making people smile. When a motivation is stated sessions away from the activity it explains, and in different words, it can fall outside the top-ranked results of direct embedding retrieval.

\section{Conclusion}

Preserving history does not ensure that later queries can recover relevant evidence through a bounded retrieval interface. CABLE addresses this problem at the association-and-retrieval layer by constructing and reusing verified antecedent links that complement direct semantic retrieval. Experiments across two benchmarks, three memory systems, and multiple LLMs show consistent overall gains.
The strongest improvements occur when useful evidence is distributed across memories or sessions, and gains on adaptive-retrieval SimpleMem and graph-based Mem0\textsuperscript{g} show that the same principle extends beyond a single host architecture.
Together, these results position retriever complementarity as a practical mechanism for improving evidence reachability in long-term memory.

\section*{Acknowledgments}
This research is supported by the Key R\&D Program of Shandong Province, China (2024CXGC010213). We express our gratitude to the funding agency for their support.

\bibliography{references}
\bibliographystyle{colm2026_conference}

\appendix
\newpage

\section{Integration protocols}
\label{app:protocols}

Table~\ref{tab:integration-protocols} summarizes how CABLE is integrated
with each host system. The comparison preserves each host system's
native retrieval regime: A-MEM and Mem0$^{g}$ use matched fixed
retrieved-entry budgets, whereas SimpleMem retains its adaptive
reflection-based protocol.

\begin{table}[H]
\centering
\small
\begin{tabularx}{\linewidth}{@{}p{0.13\linewidth}
                               p{0.24\linewidth}
                               p{0.27\linewidth}
                               X@{}}
\toprule
\textbf{Host}
&
\textbf{Host retrieval}
&
\textbf{CABLE activation}
&
\textbf{Budget control}
\\
\midrule

A-MEM
&
Direct host retrieval
&
CABLE expands after host retrieval and replaces entries only when
verified, novel candidates are available.
&
Both systems pass at most 45 entries to the answer generator. Up to five
CABLE candidates replace the five lowest-ranked baseline entries.
\\

\midrule

SimpleMem
&
Host-controlled adaptive retrieval with an LLM reflection step
&
CABLE is activated only when the reflection step judges the baseline
retrieval insufficient.
&
SimpleMem retains its native adaptive protocol; no fixed
cross-system entry budget is imposed.
\\

\midrule

Mem0$^{g}$
&
Graph-enabled host retrieval over extracted entities and relations
&
CABLE expands after the host graph retrieval.
&
Both systems use a fixed 20-entry budget. Up to five CABLE candidates
replace baseline entries.
\\

\bottomrule
\end{tabularx}

\caption{Integration protocols preserve each host system's retrieval
regime. A-MEM and Mem0$^{g}$ use matched fixed retrieved-entry budgets,
whereas SimpleMem retains adaptive retrieval and invokes CABLE only after
insufficient baseline retrieval.}
\label{tab:integration-protocols}
\end{table}

\section{Construction Cost, Graph Growth, and Retrieval-Time Overhead}
\label{app:overhead}

CABLE introduces additional LLM calls during memory-link construction,
but adds no LLM calls at retrieval time.

\paragraph{Construction cost.}
When a new memory $m_i$ arrives, CABLE performs one type-classification
call, one query-generation call, and one verification call per surviving
candidate in $C_i$. Link construction is incremental: CABLE forms edges
only between $m_i$ and its retrieved candidates and does not recompute
the graph over previously stored memories. The LLM-call cost per memory
is therefore bounded by the number of candidates that survive overlap
subtraction, rather than by the size of the memory base. This cost is paid
once per write and amortized over subsequent reads.

\paragraph{Graph growth.}
Overlap subtraction discards candidates already inside the host
retriever's direct top-$K_b$ neighborhood, while verification rejects
relations based only on topical co-occurrence or entity overlap and pairs
containing non-substantive memories. CABLE therefore avoids all-pairs
verification. Because each new memory generates at most $N_q$
antecedent-oriented queries and each query retrieves at most $K_h$
candidates, CABLE considers at most $N_qK_h$ candidate antecedents and
stores at most $N_qK_h$ new edges per memory. Under fixed
hyperparameters, the total number of stored edges therefore grows at
most linearly with the number of memories:
\[
|E| = O(|M|).
\]

\paragraph{Retrieval-time overhead.}
At query time, CABLE adds no LLM calls. For the selected seed set $S$, it
enumerates one-hop graph neighborhoods, scores candidates using
available embedding similarities, and applies novelty filtering. Before
deduplication and candidate scoring, neighbor enumeration is
proportional to
\[
\sum_{s\in S}\deg(s).
\]
The global bound $|E|=O(|M|)$ does not imply that every seed has constant
degree. In the A-MEM and Mem0$^{g}$ integrations, the retrieved-entry
count remains fixed; SimpleMem retains its host-controlled adaptive
protocol.

\section{Component and Query-Design Ablations}
\label{app:ablations}
Table~\ref{tab:ablation} reports the component and query-design ablations on A-MEM with Qwen3.5-27B. The component ablations are evaluated on both benchmarks, whereas the generic query-decomposition comparison is available only on LoCoMo.

Removing overlap subtraction reduces performance by 0.91 percentage points on LoCoMo and 0.67 points on MA-LongMemEval, supporting the use of retriever-complementary rather than redundant links. Removing verification produces a smaller reduction on LoCoMo but a substantially larger 2.66-point reduction on MA-LongMemEval, where longer histories can introduce noisier antecedent candidates. Replacing the type-conditioned antecedent queries with generic topical decomposition reduces the LoCoMo score by 0.46 points, indicating a modest additional benefit from explicitly targeting prior causes, motivations, plans, and background events.

\begin{table}[t]
\centering
\small
\setlength{\tabcolsep}{6pt}
\begin{tabular}{lrrrr}
\toprule
Variant & LoCoMo & $\Delta$ & MA-LongMemEval & $\Delta$ \\
\midrule
Full CABLE
    & \textbf{74.81} & -- & \textbf{65.33} & -- \\
w/o overlap subtraction
    & 73.90 & $-0.91$ & 64.66 & $-0.67$ \\
w/o verification
    & 74.61 & $-0.20$ & 62.67 & $-2.66$ \\
Generic query decomposition
    & 74.35 & $-0.46$ & -- & -- \\
\bottomrule
\end{tabular}
\caption{Removing overlap subtraction lowers performance on both benchmarks, while removing verification causes the largest degradation on MA-LongMemEval. Generic query decomposition also underperforms the type-conditioned antecedent queries on LoCoMo. Results use A-MEM with Qwen3.5-27B; scores are mean LLM-judge percentages, and $\Delta$ is relative to full CABLE in percentage points. A dash denotes an unevaluated setting.}
\label{tab:ablation}
\end{table}

\begin{table}[t]
\centering
\setlength{\tabcolsep}{4pt}
\renewcommand{\arraystretch}{1.06}
\begin{tabularx}{\linewidth}{>{\raggedright\arraybackslash}p{0.18\linewidth} >{\raggedright\arraybackslash}X}
\toprule
\textbf{Item} & \textbf{Evidence} \\
\midrule
\textbf{Question} & Why did Melanie choose to use colors and patterns in her pottery project? \\
\textbf{Ground truth} & She wanted to catch the eye and make people smile. \\
\midrule
\textbf{Direct-retrieval seed} & \texttt{D5:6}: ``I'm a big fan of pottery -- the creativity and skill is awesome. Plus, making it is so calming. Look at this!'' \\
\midrule
\textbf{Stored CABLE link}
&
D5:6 $\rightarrow$ D12:6
\\
\midrule
\textbf{CABLE-expanded evidence} & \texttt{D12:6}: ``Thanks, Caroline! I'm obsessed with those, so I made something to catch the eye and make people smile. Plus, painting helps me express my feelings and be creative. Each stroke carries a part of me.'' \\
\midrule
\textbf{Answers} & Baseline: ``Not mentioned in context.'' \newline
+CABLE: ``To catch the eye, make people smile, and express her feelings creatively.'' \\
\bottomrule
\end{tabularx}
\caption{A concrete evidence-reachability failure and its recovery.
Direct retrieval returns a topically similar pottery memory that lacks
the required motivation and yields an abstaining answer. CABLE expands
through a stored antecedent link, recovers the otherwise unretrieved
motivation in D12:6, and produces a reference-supported answer.}
\label{tab:case_study}
\end{table}

\section{Qualitative case study}
\label{app:case_study}

Table~\ref{tab:case_study} provides the full qualitative example discussed in Section~\ref{sec:case_study}.
It shows the user question, the baseline retrieved memory, the CABLE-expanded memory, and why the CABLE memory supplies the missing explanatory evidence.

The baseline memory is close to the query because both mention pottery, making it a natural semantic match. However, it only establishes that Melanie enjoys pottery and does not explain why she used colors and patterns. The CABLE-expanded memory is absent from the baseline result set but contains the reference motivation: Melanie wanted the colors and patterns to catch the eye and make people smile. This illustrates the intended role of CABLE’s antecedent links: they do not replace semantic retrieval, but add complementary evidence when direct similarity retrieves the topic without the explanation.

\section{Prompts}
\label{app:prompts}

We list the prompts used by CABLE. Text in \textcolor{pbSlot}{\ttfamily
\bfseries braces} marks a slot filled at run time or an expected output
schema. \texttt{content} is the memory text, and
\texttt{persons} is the participant list recorded in the memory metadata, or
\texttt{Unknown} when unavailable. Colors group the prompts by pipeline
stage: memory typing, antecedent-query generation, link verification, and
evaluation. For readability, Markdown emphasis markers in the original prompts are
rendered typographically; all other prompt text is reproduced verbatim.

\subsection{Memory type classification}
\label{app:prompt_type}

Each memory is first assigned an operational type, which selects the
antecedent-query template used in the next step. Decoding uses temperature
$0.1$. If the response cannot be parsed into one of the four types, CABLE
falls back to a keyword-based classifier.

\begin{promptbox}[pbClassify]{Memory type classification}
Classify the following memory content into exactly ONE type.

Content: {content}

Types:
- opinion: expresses a personal view, preference, or judgment
- event: describes something that happened
- plan: describes future intentions or plans
- state_change: describes a change in status, condition, or relationship

Return JSON: {(*@\pbquote@*)type(*@\pbquote@*): (*@\pbquote@*)opinion(*@\pbquote@*) | (*@\pbquote@*)event(*@\pbquote@*) | (*@\pbquote@*)plan(*@\pbquote@*) | (*@\pbquote@*)state_change(*@\pbquote@*)}
\end{promptbox}

\subsection{Antecedent-oriented query generation}
\label{app:prompt_query}

Guided by the assigned type, CABLE generates at most $N_q=3$
antecedent-oriented queries. Decoding uses temperature $0.2$. The four
type-specific templates and the fallback template are shown below.
If the type-classification output cannot be parsed, CABLE applies a
keyword-based classifier. If neither the LLM output nor the rule-based
classifier yields a supported type, CABLE uses the fallback query
template in Prompt~\ref{fallback}.

\Needspace{18\baselineskip}
\begin{samepage}
\begin{promptbox}{Antecedent queries --- \texttt{event}}
An EVENT happened. Find PRIOR MEMORIES about:
EVENT: {content}
PERSONS: {persons}

Focus on:
1. Preceding events - What led up to this?
2. Goals or plans - Was this planned?
3. Related past experiences

Generate 2-3 search queries. DO NOT fabricate.
\end{promptbox}
\end{samepage}

\begin{promptbox}{Antecedent queries --- \texttt{opinion}}
A person expressed an OPINION. Find PRIOR MEMORIES about:
OPINION: {content}
PERSONS: {persons}

Focus on:
1. Experiences that shaped this opinion
2. Previous statements on related topics
3. Personal history related to this topic

Generate 2-3 search queries. DO NOT fabricate.
\end{promptbox}

\begin{promptbox}{Antecedent queries --- \texttt{plan}}
A PLAN was expressed. Find PRIOR MEMORIES about:
PLAN: {content}
PERSONS: {persons}

Focus on:
1. Motivations for this plan
2. Related past experiences
3. Expressed interests or goals

Generate 2-3 search queries. DO NOT fabricate.
\end{promptbox}

\begin{promptbox}{Antecedent queries --- \texttt{state\_change}}
A STATE CHANGE occurred. Find PRIOR MEMORIES about:
STATE CHANGE: {content}
PERSONS: {persons}

Focus on:
1. Previous state before this change
2. Triggers or causes
3. Related developments

Generate 2-3 search queries. DO NOT fabricate.
\end{promptbox}

\begin{promptbox}{Antecedent queries --- fallback}[fallback]
Find PRIOR MEMORIES for:
MEMORY: {content}
PERSONS: {persons}

Focus on causes, motivations, and background context.
Generate 2-3 search queries. DO NOT fabricate.
\end{promptbox}

\subsection{Link verification}
\label{app:prompt_verify}

Every candidate in $C_i$ is verified before an edge is inserted. Decoding
uses temperature $0.0$.

\begin{promptbox}[pbVerify]{Link verification}
Determine if the HISTORICAL memory could serve as USEFUL BACKGROUND or IMPLICIT EVIDENCE for understanding the CURRENT memory.

CURRENT MEMORY: {content_a}
HISTORICAL MEMORY: {content_b}

FIRST: If EITHER memory is non-substantive (a greeting, short acknowledgment,
    filler, simple question with no information, or generic reaction), return {(*@\pbquote@*)valid(*@\pbquote@*): false, (*@\pbquote@*)reason(*@\pbquote@*): (*@\pbquote@*)non-substantive content(*@\pbquote@*)}.

We are looking for NON-OBVIOUS connections. Surface similarity is NOT required.

VALID connections:
1. Distant causes (B led to A, even through multiple steps)
2. Background context (B helps explain WHY A happened)
3. Character development (B shows earlier state that evolved into A)
4. Implicit motivation (B reveals goals/values explaining A)
5. Multi-hop evidence (B could help answer questions about A)

INVALID connections:
1. Different people with NO interaction or influence
2. Pure coincidence (same time but unrelated)
3. No plausible reasoning chain from B to A

Return JSON: {(*@\pbquote@*)valid(*@\pbquote@*): true/false, (*@\pbquote@*)reason(*@\pbquote@*): (*@\pbquote@*)one sentence explanation(*@\pbquote@*)}
\end{promptbox}

\subsection{LLM-as-judge}
\label{app:prompt_judge}

The judge receives the question, the reference answer, and the generated
answer, and returns a score together with a short justification. Although the grading logic specifies the two endpoint values 1.0 and 0.0, the judge occasionally returns an intermediate value. We retain the returned score without rounding or binarization. Accordingly, all reported results are mean LLM-judge scores, expressed as percentages. Baseline and
+CABLE systems are evaluated with this same prompt and the same judge model
within each benchmark-model setting.

\begin{promptbox}[pbJudge]{LLM-as-judge}
You are an expert Relevance & Accuracy Evaluator. Your task is to determine if the Predicted Answer successfully retrieves the necessary information to answer the Question, based on the Reference Answer.

Question: {question}
Reference Answer: {reference}
Predicted Answer: {prediction}

Evaluation Criteria:

1. (*@\textbf{Responsiveness to Query}@*):
   The predicted answer must directly address the specific question asked. It must contain highly relevant information that is topically aligned with the user's intent.

2. (*@\textbf{Core Fact Preservation}@*):
   The prediction must capture the (*@\pbquote@*)Key Signal(*@\pbquote@*) or (*@\pbquote@*)Core Entity(*@\pbquote@*) from the reference. The primary subject (Who), event (What), or outcome must be factually grounded in the reference text.

3. (*@\textbf{Informational Utility}@*):
   The answer must provide actionable or meaningful value. Even if brief, it must convey the essential message required by the question context.

4. (*@\textbf{Acceptable Representational Variances (Robustness Protocol)}@*):
   To ensure fair evaluation of semantic meaning over syntactic rigidity, you must accept the following variations as (*@\textbf{Valid Matches}@*):
   - (*@\textbf{Temporal \& Numerical Margins}@*): Accept timestamps within a reasonable proximity (e.g., +/- 1-2 days due to timezone/reporting differences) and rounded numerical approximations.
   - (*@\textbf{Granularity Independence}@*): Accept answers at different levels of abstraction (e.g., (*@\pbquote@*)Afternoon(*@\pbquote@*) vs. (*@\pbquote@*)14:05(*@\pbquote@*), (*@\pbquote@*)Late October(*@\pbquote@*) vs. (*@\pbquote@*)Oct 25th(*@\pbquote@*)) provided they encompass the truth.
   - (*@\textbf{Information Subsetting}@*): A valid subset of the reference (e.g., mentioning 1 out of 3 reasons) is acceptable if it answers the core of the question.
   - (*@\textbf{Synonymy}@*): Recognize domain-specific synonyms and different formats as equivalent.

Grading Logic:
- Score 1.0 (Pass): The prediction contains relevant core information, answers the question with sufficient utility, OR falls within the acceptable representational variances defined in criterion #4.
- Score 0.0 (Fail): The prediction contains NO relevant information, fails to identify the core subject/event, or provides no key info that matches the question's intent.

{
  (*@\pbquote@*)score(*@\pbquote@*): 1.0,
  (*@\pbquote@*)reasoning(*@\pbquote@*): (*@\pbquote@*)Brief assessment focusing on information relevance and core match.(*@\pbquote@*)
}

Return ONLY the JSON, no other text.
\end{promptbox}

\section{Limitations}
\label{app:limitations}

CABLE focuses on constructing and reusing memory associations that extend
direct retrieval, rather than redesigning the full memory lifecycle. As a
plug-in augmentation, it leaves how memories are written, updated,
consolidated, and discarded to the host system. In the current implementation,
link construction is append-only: each verified association becomes a
persistent edge, and no mechanism removes or revises edges after insertion.
The graph can therefore continue to grow over an agent's lifetime, even though
overlap subtraction and verification limit which edges are initially added.

This has two consequences. First, retrieval-time expansion visits the one-hop neighborhoods of the selected seeds, so its cost depends on their degrees and may increase as associations accumulate. Second, an edge can continue to surface
evidence after a later memory has superseded or corrected it. Addressing these
issues will require forgetting, decay, or consolidation policies that determine
which associations should remain active. Integrating such policies with CABLE
is an important direction for future work.

Our experiments use long-term conversational memory question answering to
evaluate whether CABLE improves access to relevant historical evidence. This
setting isolates the retrieval problem studied here, but does not cover
tool-using agents, multi-agent coordination, or graph-level task execution.
Evaluating CABLE in these broader agent workflows remains an important
direction for future work.

\section{Use of LLMs}
We use LLMs for memory extraction, CABLE link construction, answer generation, and LLM-as-judge evaluation. LLMs were also used to assist with manuscript framing, restructuring, and language revision. All LLM-generated suggestions were critically reviewed, verified, and revised by the authors, who take responsibility for all reported results and manuscript content.

\end{document}